\documentclass[final]{cvpr}

\usepackage{times}
\usepackage{epsfig}
\usepackage{graphicx}
\usepackage{amsmath}
\usepackage{amssymb}
\usepackage{color}
\usepackage{bm}
\usepackage{multirow}
\usepackage{diagbox}
\usepackage{array}
\usepackage{booktabs}
\usepackage{nth}
\usepackage{nicefrac}
\usepackage{relsize}
\usepackage{xspace} 
\usepackage{bm}
\usepackage{enumerate}
\usepackage[table]{xcolor}
\usepackage{enumitem}
\usepackage{array}

\usepackage{cuted}
\usepackage{capt-of}
\newcolumntype{P}[1]{>{\centering\arraybackslash}p{#1}}
\newcommand{\mytilde}{\raise.17ex\hbox{$\scriptstyle\mathtt{\sim}$}}
\graphicspath{ {./images/} }
\usepackage{relsize,bbm, balance}

\usepackage[pagebackref=false,breaklinks=true,colorlinks,bookmarks=false]{hyperref}

\def\cvprPaperID{4642} % *** Enter the CVPR Paper ID here
\def\confYear{CVPR 2021}
\begin{document}

\title{Semi-supervised Synthesis of High-Resolution Editable Textures for 3D Humans}

\author{Bindita Chaudhuri$^{1*}$, Nikolaos Sarafianos$^2$, Linda Shapiro$^1$, Tony Tung$^2$\\
$^1$University of Washington, $^2$Facebook Reality Labs Research, Sausalito\\
$^1${\tt\small \{bindita,shapiro\}@cs.washington.edu}, $^2${\tt\small \{nsarafianos,tony.tung\}@fb.com}}

% \maketitle

\twocolumn[{%
\renewcommand\twocolumn[1][]{#1}%
\maketitle
\vspace{-10pt}
\centering
\includegraphics[width=0.975\linewidth]{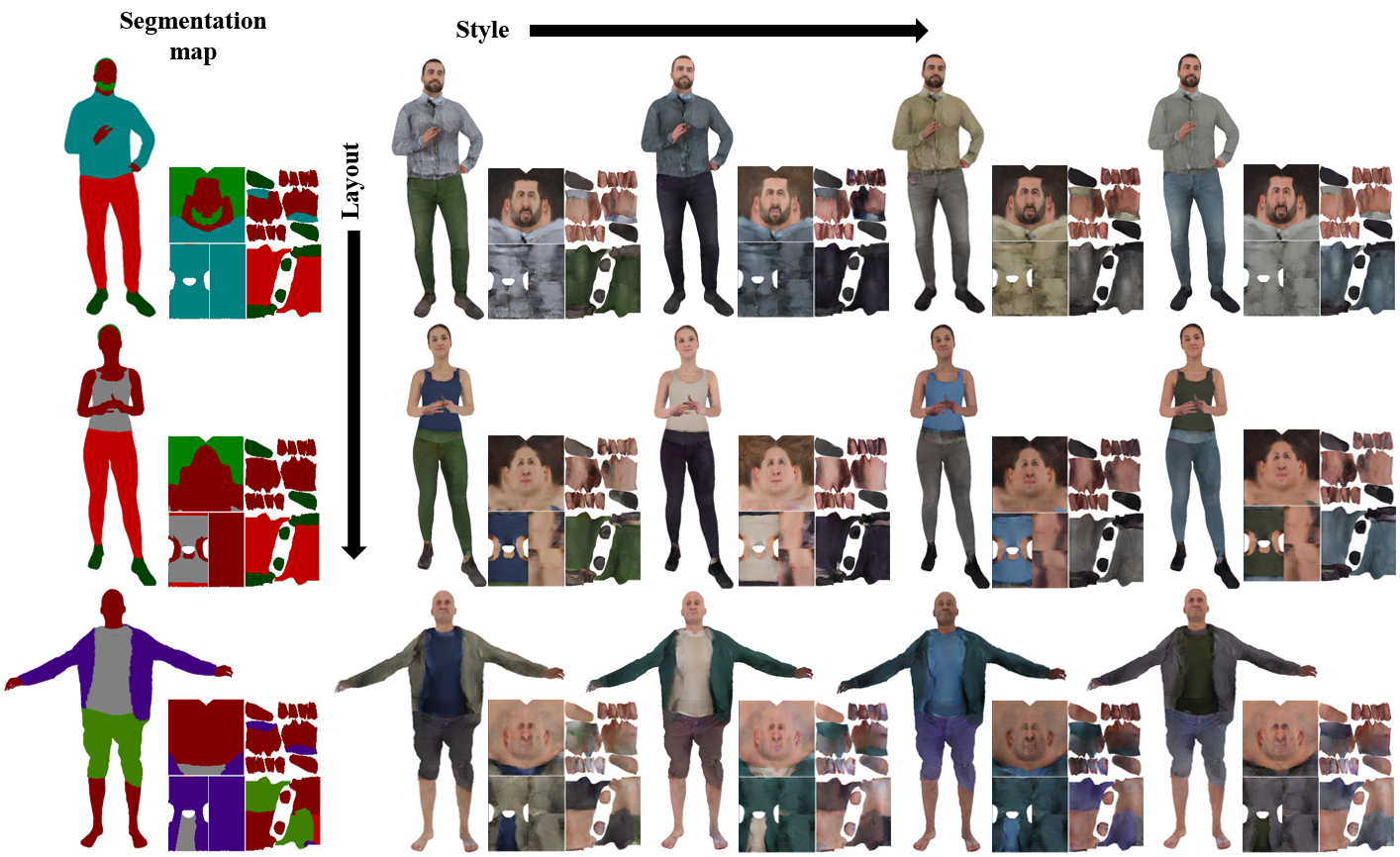}
\captionof{figure}{Given a segmentation map defining the layout of semantic regions in a texture map, our proposed method generates diverse high-resolution texture maps which are then used to render 3D humans. Each example shows a UV map in the inset and the corresponding 3D mesh rendered with the map. The style of each region/class can be controlled individually by manipulating the input style vectors. Note that along each column, the styles of the same classes are the same (for example, the man in the first row and the woman in the second row of the first column are wearing green colored pants).}
\label{fig:teaser}
\vspace{12pt}
}]

\begin{abstract}
     We introduce a novel approach to generate diverse high fidelity texture maps for 3D human meshes in a semi-supervised setup. Given a segmentation mask defining the layout of the semantic regions in the texture map, our network generates high-resolution textures with a variety of styles, that are then used for rendering purposes. To accomplish this task, we propose a Region-adaptive Adversarial Variational AutoEncoder (ReAVAE) that learns the probability distribution of the style of each region individually so that the style of the generated texture can be controlled by sampling from the region-specific distributions.\let\thefootnote\relax\footnote{{$^*$This work was conducted during an internship at FRL Research.}} In addition, we introduce a data generation technique to augment our training set with data lifted from single-view RGB inputs. Our training strategy allows the mixing of reference image styles with arbitrary styles for different regions, a property which can be valuable for virtual try-on AR/VR applications. Experimental results show that our method synthesizes better texture maps compared to prior work while enabling independent layout and style controllability.
\end{abstract}

\section{Introduction}

3D human avatar creation has recently gained popularity with the growing use of AR/VR devices and virtual communication. A human body is represented by a 3D surface mesh modeling its shape, and a texture map (an image in UV space) encoding its appearance mapped to the 3D surface. Realistic textures for avatars are crucial for more immersive experiences with believable digital humans. To date, it is still tedious to create texture maps as it may require hours of manual work by a technical artist or special equipment (e.g., 3D scans, multiview-camera setting, etc.) to capture all the body and cloth details. Hence in this work, we develop a novel method to synthesize photorealistic texture maps for human 3D meshes in a semi-supervised setup with the following properties: i) high resolution, ii) high fidelity, iii) large diversity, and iv) editability.

Recent deep learning-based techniques for textured 3D human generation~\cite{lahner2018deepwrinkles, CVPRarch, 360DegreeTO} infer the textures from 2D clothed human images, which cause their textures to be limited to the garment styles in the image dataset. The fidelity of the inferred textures is also constrained by the resolution of the 2D images. Prior work~\cite{Lassner:GeneratingPeople:2017, men2020controllable, MISC} relies on image-to-image translation networks to convert a human body part segmentation mask into a textured image. These techniques directly generate a clothed human image instead of a texture image that can be applied to a 3D mesh. Besides, their style controllability is limited to mostly changing garment colors but not the actual styles like floral or checkered patterns. Among the unsupervised image synthesis works, StyleGAN~\cite{stylegan} and StyleGAN2~\cite{stylegan2} can generate high-resolution and high-fidelity results with their unconditional image synthesis setup, but such a setup does not allow easy controllability for texture maps that come with a predefined layout in the UV space. Conditional image synthesis techniques like Pix2PixHD~\cite{pix2pixHD} and SPADE~\cite{SPADE} use a conditional GAN to associate each input segmentation mask to a unique output image. While the VAE version of SPADE introduces some controllability, it can only control the global style but not class-specific styles of the output image. The authors of SEAN~\cite{SEAN} overcame this problem by encoding class-specific styles that are then used to learn the normalization parameters for the conditional GAN. This allows them to apply different styles to different regions using different exemplar images, one per region. As a result, exemplar-based approaches are limited to reconstructing the existing textures or linearly interpolating between them. Besides, it is difficult and time-consuming to find several different exemplar images for different styles.

To address these issues, we propose a novel architecture that we call Region-adaptive Adversarial Variational AutoEncoder (ReAVAE) that learns the probability distributions of per-region styles from texture maps using a VAE in a semi-supervised setup and allows per-region style controllability of the output texture using the learned distributions. Our architecture has three components. First, the style encoder encodes an input texture map and performs region-wise average pooling of the encoded features based on the semantic segmentation mask corresponding to the input texture to produce per-class feature vectors. Second, the VAE bottleneck learns to approximate the features of each class by a standard normal distribution, from which a random sample is generated to produce a transformed feature vector. Lastly, the generator takes the per-class transformed feature vectors, a segmentation mask, and random Gaussian noise as inputs to generate the desired texture map. The generated map is then converted to higher resolution by passing it through a pretrained image super-resolution network and finally rendered using a differentiable renderer. During inference, we solely use the generator that enables independent layout controllability through the input mask and per-region style controllability through the input random vectors, which results in the generation of a wide variety of textures.
We also introduce a training strategy that enables our network to perform both reconstruction of an input image and generation of an arbitrary image. Hence, we can mix the styles of some regions of the input image with arbitrary styles for the remaining regions by manipulating the input per-region feature vectors of the generator. Finally, to alleviate the problem of having limited data originating from textures from 3D scans, we introduce a method to generate training data for our network by lifting textures from full-body clothed human images to the UV space. In summary, our contributions are:
\begin{enumerate}
\itemsep0em 
    \item We propose a novel architecture for semi-supervised synthesis of diverse high-fidelity texture maps for 3D humans, given the layouts (segmentation masks) as input, with independent layout and style controllability. The textures can be used for high-resolution rendering. To the best of our knowledge, no existing work has tackled this task to date.
    \item We utilize a VAE to learn the distributions of the styles of each region separately, thereby allowing the user to sample from region-specific distributions during inference to generate a variety of textures. Our training scheme allows mixing styles from exemplar images for some regions with arbitrary styles for other regions, a useful property for 3D virtual try-on applications.
    \item We introduce a training data generation technique that lifts textures from single-view RGB images of full-body clothed humans to the UV space.
\end{enumerate}

\section{Related Work}

\noindent\textbf{Image synthesis}: Among the recent works on unsupervised data generation using Generative Adversarial Networks (GANs)~\cite{GAN}, papers such as the Progressive GAN \cite{progressivegan}, StyleGAN \cite{stylegan}, and StyleGAN2 \cite{stylegan2} can generate high resolution and high fidelity images. Since the diversity of the generated images is directly proportional to the size of the training data, a recent line of works has proposed techniques that can generate considerable diversity with limited data. This is done by effectively fine-tuning a pretrained StyleGAN2 network~\cite{minegan} or by applying differential augmentation~\cite{karras2020training,diffaugment} at the generator outputs.

However, for our texture map generation, the latent vectors of StyleGAN2 encode both the layout and the global styles together in a complicated manner which does not allow for easy editability. Given a segmentation map as input, Pix2PixHD~\cite{pix2pixHD} used an image-to-image translation~\cite{choi2018stargan,huang2018multimodal, lee2018diverse, liu2019few} method to generate output image and SPADE \cite{SPADE} improved upon Pix2PixHD by using the segmentation map in the spatially-adaptive normalization layers. To overcome their limitation of allowing no or only global style controllability, SEAN~\cite{SEAN} introduced semantic-region adaptive normalization to add class-specific style controllability to conditional image synthesis. However, SEAN relies on one or more exemplar images for the style transfer and hence cannot be used in our desired non-exemplar based setup.

\noindent\textbf{Textured human image synthesis}: Recent works on clothed human image generation usually perform garment transfer using exemplar-based conditional image synthesis technique. For example, \cite{texturegan} uses a full-body sketch as a condition and a texture patch image as an exemplar, \cite{neuralrenderingofhumans, exampleguided, crossdomain} use a 2D human pose image~\cite{sarafianos20163d} as a condition and clothed human images as exemplars. Non-exemplar based methods include~\cite{Lassner:GeneratingPeople:2017, MISC}, which learn to generate textured human images given input segmentation masks via image-to-image translation. However, these methods generate low-resolution textured humans directly in the 2D space, and hence cannot be utilized on 3D human meshes. Methods which generate 3D textured humans in clothing~\cite{Alldieck_2019_CVPR, multigarmentnet, CVPRarch, 360DegreeTO, pifuhd, humanparsing} are mainly focused on reconstructing the 3D geometry from one~\cite{CVPRarch, pifuhd, humanparsing} or more~\cite{Alldieck_2019_CVPR,multigarmentnet} RGB images with the texture colors embedded as vertex information in the geometry. While the work of Lazova \etal~\cite{360DegreeTO} generates a texture map as an intermediate step, the method is reconstruction-based only. Other works such as~\cite{clothflow, Vibe, clothingdatasettransfer,deepfashion3D} generate 3D garment textures from a dataset of 2D garment RGB images either by using garment templates or by using body shape and pose as reference.
\begin{figure}[t]
    \centering
    \includegraphics[width=0.975\linewidth]{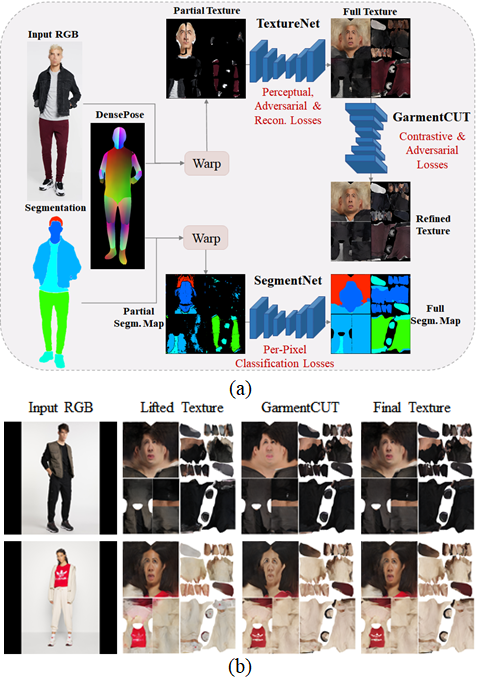}
    \caption{\textbf{Training data generation}. (a) Pipeline showing lifting the data from RGB inputs and refining the textures using GarmentCUT, (b) results of lifting and refining.}
    \label{fig:cutprocess}
\end{figure}

\section{Methodology}

\begin{figure*}
    \centering
    \includegraphics[width=1.0\linewidth]{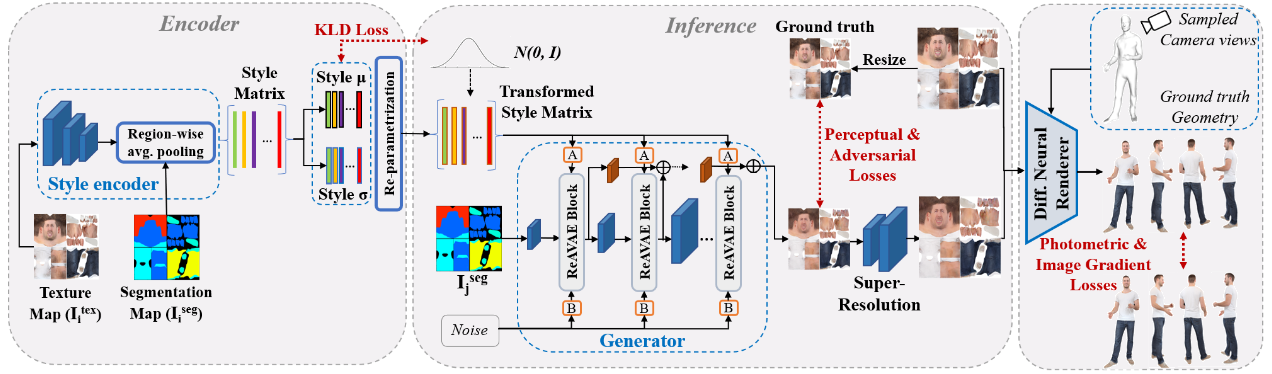}
    \caption{\textbf{Our end-to-end framework}. The style encoder encodes the region-wise styles from the input which are then used by our ReAVAE to learn region-specific style distributions. The generator synthesizes texture maps given per-class style vectors and desired layout (segmentation map). The generated texture is then upscaled and used to render 3D human meshes.}
    \label{fig:architecture}
\end{figure*}

\subsection{Training Data Generation}

\noindent\textbf{Dataset of registered scans} Our ground truth training data is composed of 500 3D scans (single texture per scan) from the RenderPeople~\cite{renderppl} dataset and 400 3D scans (five textures per scan) from the AXYZ~\cite{axyz} dataset, resulting in 2300 textures for training and 200 for testing. The scans are watertight meshes, with the subjects wearing a wide variety of clothes and holding different types of objects such as backpacks and phones. To obtain the texture maps in the UV space, we performed non-rigid registration of a body template similar to SMPL~\cite{loper2015smpl} to the scans along with additional 2D landmark constraints in order to handle complicated poses. To obtain the segmentation maps we first rendered all scans using Blender Cycles~\cite{blender} from 180 different viewpoints and then ran a state-of-the-art cloth segmentation algorithm~\cite{fu2019imp} to obtain instance segmentations in the image space. The instances comprised of hair, skin, a variety of different garments, and a few accessories for a total of 28 classes. We then lift all the segmentation estimates from the RGB to the UV space using the method of Lazova \etal~\cite{360DegreeTO} and aggregate their results by selecting the most frequently predicted class across all views for each pixel. Finally, we merged the classes that were semantically similar (\eg, jacket with hoodie) for a total of 20 distinct classes.

\noindent\textbf{Data lifting} When we set up our baselines we quickly observed that the amount of data we had was quite small to learn the distribution of each class, with lack of diversity in terms of identity and garment styles. To overcome this limitation, we introduce a novel approach that lifts textures from single-view RGB images of full-body clothed humans to the UV space inspired by a recent work~\cite{360DegreeTO}. Specifically, we first run DensePose \cite{densepose} on the RGB image to obtain IUV estimates in the image space which are then used to lift the input RGB image to the UV space to obtain a partial texture map, which is then passed through a pretrained neural network to generate the complete texture map. Similarly, the cloth segmentations were first lifted to the UV space and then completed using a pretrained neural network to obtain the complete segmentation maps.

\noindent\textbf{Unpaired data refining} As one would expect, our lifting process generates textures that are noisier than the ones obtained from the registered scans, with artifacts on the occluded portions and baked lighting on the skin and clothes. To address these shortcomings we propose to use CUT~\cite{cut}, an unpaired image-to-image translation method that aims to maximize the mutual information between images of two different domains. Given a dataset $X$ containing the registered scan textures and a dataset $Y$ containing the lifted textures from the RGB images, our network, which we call GarmentCUT, samples unpaired instances and learns the mapping from $Y$ to $X$. To train this network we use the adversarial GAN loss and the patchwise contrastive loss together with the hyperparameters as used in \cite{cut}. Finally, while for the clothed regions this approach worked remarkably well, this was not the case for the face region. We attribute this outcome to the fact that in the textures obtained from the registered 3D scans, there is a limited number of unique identities that ended up affecting the unpaired translation training to change the identity of the subject to some extent which is not desirable. Hence we propose to keep the area that corresponds to the face from the lifted RGB textures and use the rest of the texture map produced by GarmentCUT. We show the complete pipeline for the training data generation in Fig.~\ref{fig:cutprocess}a and examples of the obtained results in Fig.~\ref{fig:cutprocess}b. We used 8,000 images from the DeepFashion dataset~\cite{liuLQWTcvpr16DeepFashion} equally sampled in terms of garments that were then processed using our proposed data generation approach to enhance our training set.

\subsection{Network Architecture}
Our network ReAVAE comprises of 3 major components: (a) style encoder, (b) VAE bottleneck and (c) generator. An overview of our method is shown in Fig.~\ref{fig:architecture}.

\noindent\textbf{Style encoder}: The style encoder encodes the style of each class \(c_k, k \in[1,C]\) into a \(W\)-length style vector $\hat{S}_c$, which together form a  $C \times W$ style matrix. Given the $i$-th texture map $I^{\text{tex}}_i$ and its corresponding segmentation map $I^{\text{seg}}_i$ as inputs, the style encoder first extracts \(W\)-length features from $I^{\text{tex}}_i$ using an encoder similar to~\cite{SEAN}. Then, for each class in $I^{\text{seg}}_i$, the feature values at the pixel locations belonging to that class are spatially averaged to obtain the style in the form of a vector. If a class is missing in $I^{\text{seg}}_i$, its feature vector is set to the zero vector.

\noindent\textbf{VAE bottleneck}: This is the main component that enables the network to learn the probability distributions of the styles of each class. Each vector $\hat{S}_c$ is passed through one fully-connected (FC) layer to generate the \(W\)-length mean, and through another FC layer to generate \(W\)-length variance. Hence we have \(C\) pairs of FC layers where each pair learns the mean and variance of the style of the corresponding class. We then use the reparameterization trick~\cite{reparameterization} to generate a random sample from the distribution represented by the learned mean and variance, which forms the transformed style vector $S_c$ for each class.

\begin{figure}[t]
    \centering
    \includegraphics[width=1.0\linewidth]{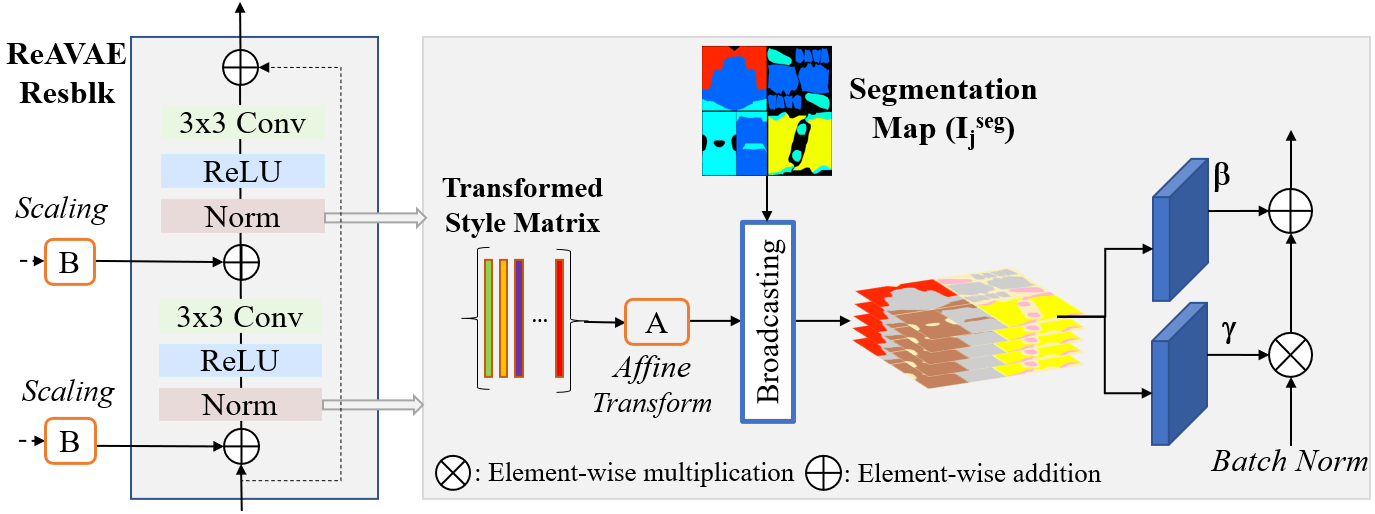}
    \caption{ReAVAE Resblk and our normalization layer.}
    \label{fig:normlayer}
\end{figure}

\noindent\textbf{Generator}: The generator (decoder of VAE) learns to synthesize the desired output texture map by taking the transformed style matrix, a guiding segmentation map $I^{\text{seg}}_j$ and Gaussian noise as inputs. It comprises multiple ResNet blocks (\ie, ReAVAE Resblks) followed by upsampling layers. We opted for a skip generator architecture~\cite{stylegan2} since it consistently outperformed other alternatives and hence, we convert the output of each Resblk into a 3-channel image using $1\times1$ convolutions and add all of them up to produce the final output. To distill the information from the transformed style matrix to the decoder of our network, we employ a normalization layer that is depicted in Fig.~\ref{fig:normlayer}. Each $S_c$ is converted to a layer-specific style vector $A(S_c)$ by passing it through an FC layer. These style codes are then broadcasted to their respective pixel locations defined by the segmentation map $I^{\text{seg}}_j$ to generate a style feature map, which is then convolved to generate the pixel-wise $\gamma$ and $\beta$ values for that layer. The addition of sampled Gaussian noise helps to learn high-frequency details~\cite{stylegan2,SEAN}.

Finally, the output is passed through a pretrained (with fixed weights) image Super-Resolution Network (SRN) to convert it to \(4\times\) its resolution and then rendered along with the ground truth 3D geometry using a differentiable renderer. We re-train a publicly available super-resolution approach~\cite{rcan} with our training data and use it as our SRN.

\subsection{Loss Functions}

\noindent\textbf{Adversarial loss}: We use the hinge loss as our adversarial loss with two multi-scale patch-based fully convolutional networks as the discriminator~\cite{SEAN} that takes the generated ($G(x)$) and ground truth ($x$) textures as input. 

\noindent\textbf{Reconstruction loss}: Instead of the pixel-wise loss which tends to generate blurry images, we use the perceptual loss for reconstruction. Specifically, we take multi-layer outputs of a pre-trained VGG~\cite{simonyan2014very} and the discriminator to compare the features of the generated and ground truth images as $L_{\text{Perc}} = \sum_{l=1}^L ||\text{VGG}_l(x) - \text{VGG}_l(G(x))||_1$ and $L_{FM} = \sum_{l=1}^3 ||\text{D}_l(x) - \text{D}_l(G(x))||_1$ respectively. The loss is given by $L_{rec} =  L_{\text{Perc}} + L_{FM}$.

\noindent\textbf{Render loss}: We render the generated and ground truth textures with the ground-truth 3D geometry from $V$ different camera viewpoints. Then, we use the per-view photometric loss $L_{ph} = ||\mathcal{R}(x) - \mathcal{R}(G(x))||_1$ and image gradient loss $L_{gr} = ||\mathcal{G}(\mathcal{R}(x)) - \mathcal{G}(\mathcal{R}(G(x)))||_1$  as our render loss defined by $L_{ren} = \frac{1}{V} \sum_{v=1}^V (L_{ph} + L_{gr})$.

\noindent\textbf{KLD loss}: We use the Kullback–Leibler divergence loss to approximate the learned style distribution for each class to a standard normal distribution \(N(0,I)\) and is formulated as $L_{KLD} = \frac{1}{2} \sum_{c=1}^{C} \sum_{w=1}^{W} (\mu_{cw}^2 + \sigma_{cw}^2 - 1 - \ln(\sigma_{cw}^2))$.

\noindent The final loss used to train our network is given by:
\begin{equation}
	\begin{aligned}
		L_{f} &= L_{adv} + \lambda_{rec} L_{rec} + \lambda_{ren} L_{ren} + \lambda_{KLD} L_{KLD}.\\
	\end{aligned}
\end{equation}

\begin{figure*}
    \centering
    \includegraphics[width=0.9\linewidth]{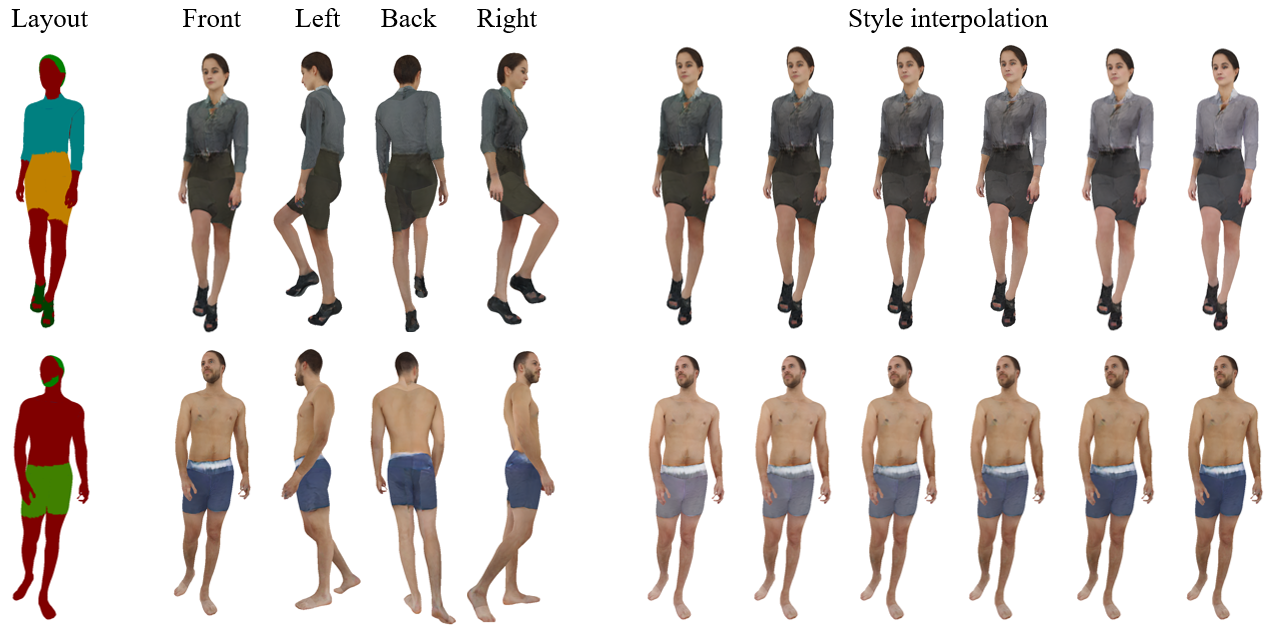}
    \caption{\textbf{Renders of 3D humans with generated textures}. Our renders are consistent across multiple viewpoints, and two random style matrices can be linearly interpolated to generate additional renders with the respective intermediate styles.}
    \label{fig:renders}
\end{figure*}

\subsection{Implementation Details}
We implement our network using Pytorch \cite{pytorch} and our differentiable renderer using Pytorch3D \cite{pytorch3d}. Our network is trained using Adam \cite{adam} optimizer ($\beta_1=0, \beta_2=0.999$) with learning rate 0.0001. The number of classes \(C\) is \(20\), number of views $V$ is 4 (front, back, left, right), the vector size \(W\) for each style vector is set to \(512\) and the weighting parameters of our loss function are set to $\lambda_{rec}=10, \lambda_{ren} = 25$ and $\lambda_{KLD} = 0.01$. Training takes about a day on a single Tesla v100 GPU with a batch size of 4. Spectral Norm~\cite{miyato2018spectral} and Synchronized Batch Norm~\cite{synchbatchnorm} are used in addition to our normalization layer. The VAE operates at \(256\times256\) images, and the final output textures and rendered images of ReAVAE are at \(1024\times 1024\) resolution.

\noindent\textbf{Training}: Since our network consists of individual components, we introduce a novel training strategy that enables our network to perform i) reconstruction of an input texture map, or ii) synthesis of a random texture map, or iii) a mixture of both. To enable reconstruction, we omit the VAE bottleneck (\ie $L_{KLD}$ from $L_f$) and directly use the style matrix as the transformed style matrix, converting the network into an autoencoder. To enable random synthesis, we use the entire pipeline provided in Fig.~\ref{fig:architecture}. We alternate between these two types of training at every iteration but in both cases, we train our framework end-to-end with the same segmentation map as input to the style encoder and the generator (\ie $I^{\text{seg}}_i = I^{\text{seg}}_j$).

\noindent\textbf{Testing}: Our network is designed in a modular manner that provides flexibility in our test setup as well. Our network can operate under four testing scenarios. The first one is reconstruction of an input texture map $I^{\text{tex}}_i$ by using the trained style encoder followed by the trained generator with $I^{\text{seg}}_i = I^{\text{seg}}_j$. The second one is style transfer between layouts, when $I^{\text{seg}}_i \neq I^{\text{seg}}_j$. The third one is the generation of a random texture map by using only the generator and giving a random layout $I^{\text{seg}}_j$ and \(C\) standard normal random vectors of length \(W\) as inputs to it. We call this the inference setup in Fig.~\ref{fig:architecture}. The fourth one is style mixing, where $\hat{S}_c$ for some classes from $I^{\text{tex}}_i$ are mixed with some random vectors for other classes. This setup will be further explored in our qualitative results described in Sec.~\ref{ssec:apps}.

\section{Results}

In this section, we evaluate the performance of our method both qualitatively and quantitatively. Fig.~\ref{fig:teaser} shows some randomly synthesized textures and the corresponding human images obtained by rendering the ground truth 3D geometry meshes with the synthesized textures. Each column has different layouts but same style (\ie the random vectors for that style are generated with the same seed), whereas each row has same layout but different styles for different classes (including skin and hair).
More examples of our rendered textures are shown in Fig.~\ref{fig:renders}. We show the renders from four camera viewpoints to demonstrate that our textures are seamless and consistent across all views. We can also interpolate between any pair of random style vectors to generate a wide variety of styles.

\subsection{Comparison with State-of-the-art Methods}

\begin{figure*}
    \centering
    \includegraphics[width=1.0\linewidth]{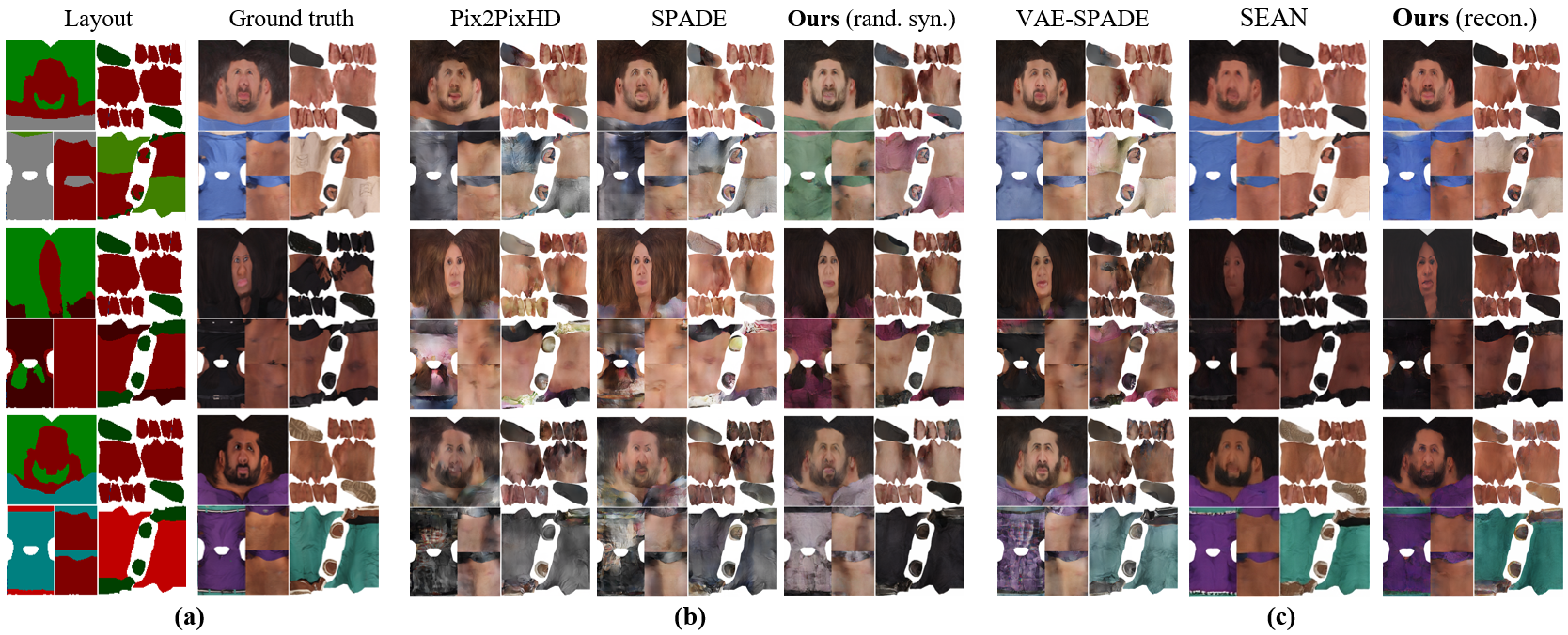}
    \caption{\textbf{Visual comparison of texture map generation}. (a) Inputs, (b) non-exemplar guided random synthesis methods, (c) exemplar-guided reconstruction methods. Our method generates results with higher fidelity, especially in the face region.}
    \label{fig:comparison}
\end{figure*}

\begin{figure}[t]
    \centering
    \includegraphics[width=1.0\linewidth]{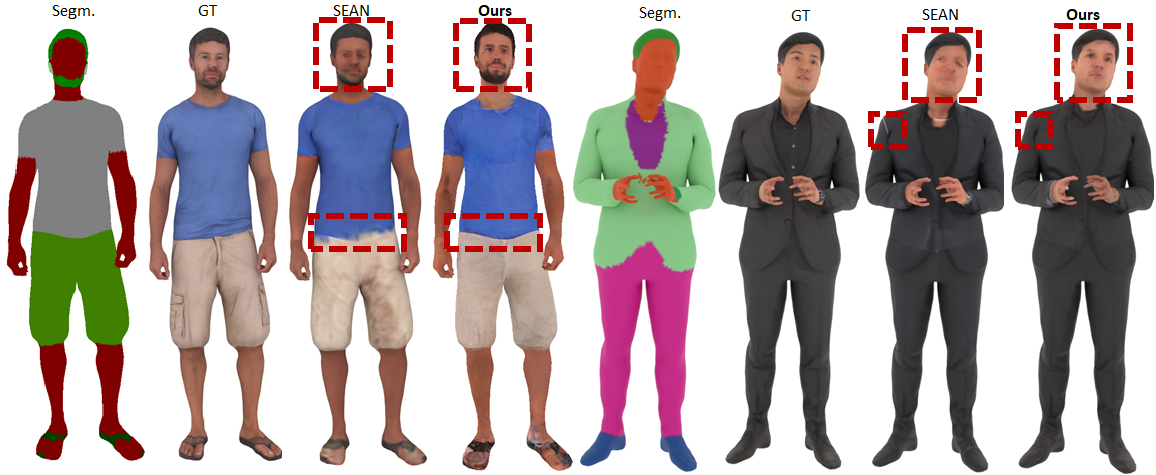}
    \caption{Qualitative comparison with SEAN~\cite{SEAN}.}
    \label{fig:comparisonWITHsean}
\end{figure}

We compare the quality of our generated textures with two categories of state-of-the-art conditional image synthesis methods: (a) non-exemplar guided random image synthesis techniques (Pix2PixHD~\cite{pix2pixHD} and SPADE~\cite{SPADE}), and (b) exemplar guided reconstruction techniques (Multimodal synthesis with SPADE (VAE-SPADE)~\cite{SPADE} and SEAN~\cite{SEAN}). All comparisons are done at 256$\times$256 resolution to be consistent across all methods.

\noindent\textbf{Quantitative evaluation}: We use the following evaluation metrics for quantitative evaluation: (a) structural similarity (SSIM) and peak signal-to-noise ratio (PSNR) for reconstruction accuracy, and (b) Fr\'echet Inception Distance (FID)~\cite{FID} and mean Kernel Inception Distance (KID)~\cite{KID} for image fidelity. Table~\ref{tab:comparison} compares the performance of our method to state-of-the-art methods. The obtained results indicate that our method clearly outperforms prior work at both reconstruction and synthesis metrics.

\setlength{\tabcolsep}{.1cm}
\begin{table}[t]
	\centering
	\caption{Quantitative comparison of our results with respect to state-of-the-art methods in terms of reconstruction accuracy (PSNR \& SSIM) and fidelity (FID \& KID).}
	\resizebox{0.9\columnwidth}{!}{
	\begin{tabular}{l c cccc}
        \toprule
        \textbf{Method} & \textbf{PSNR}\(\uparrow\) & \textbf{SSIM}\(\uparrow\) & \textbf{FID}\(\downarrow\) & \textbf{KID}\(\downarrow\)\\
        \midrule
        Pix2PixHD~\cite{pix2pixHD}  & - & - & 46.19 & 0.045\\
        SPADE~\cite{SPADE}          & - & - & 42.89 & 0.037\\
        VAE-SPADE~\cite{SPADE}      & 16.13 & 0.66 & 35.17 & 0.028 \\
        SEAN~\cite{SEAN}            & 18.92 & 0.74 & 32.45 & 0.021 \\
        \textbf{Ours}           & \textbf{19.67} & \textbf{0.79} & \textbf{29.54} & \textbf{0.015} \\
        \bottomrule  
    \end{tabular}
    \label{tab:comparison}
    }
\end{table}

\noindent\textbf{Qualitative comparison}: Fig.~\ref{fig:comparison} visually compares the textures generated by different methods. We can see that without exemplar images, Pix2PixHD and SPADE face difficulty in associating the input layout with appropriate textures. We also observed that exemplar guided techniques tend to overfit on the training data, which is the reason behind the poor quality of their textures on test data as in Fig.~\ref{fig:comparison}. More comparisons of our rendered textures with SEAN~\cite{SEAN} are shown in Fig.~\ref{fig:comparisonWITHsean} (our improvements are highlighted in red). Our network together with our training strategy ensure that we learn meaningful style features that can then be broadcasted easily to any layout.

\noindent\textbf{Limitations}: We found that \mytilde15\% of our training data contains patterns/logos, hence our learned distributions are dominated by solid colors and we have to sample several times to generate patterns which is easy to do with our designed UI. Logos, especially small ones, are challenging to generate since they tend to be distorted during the 2D to UV lifting. However, our method handles well multi-garment textures like jacket and inner-shirt separately. Also the human identity, being embedded within the style vector of the skin class, is sensitive to the layout of the generated texture. To address this, we experimented with adding face parsing masks but their impact was insignificant. Future work can investigate ways to explicitly handle the identity by potentially using a face recognition network or by collecting more diverse data. We will also explore the possibility of automatic segmentation and mask replacement as an intermediate step in our method in order to lift the constraint of depending on input segmentation masks. However, since virtual try-on applications are generally limited to few (mostly frontal) views, we assume that existing and future cloth segmentation methods perform well or the masks can be easily edited manually in case of inaccuracies.

\subsection{Ablation Study}
\begin{figure}[t]
    \centering
    \includegraphics[width=1.0\linewidth]{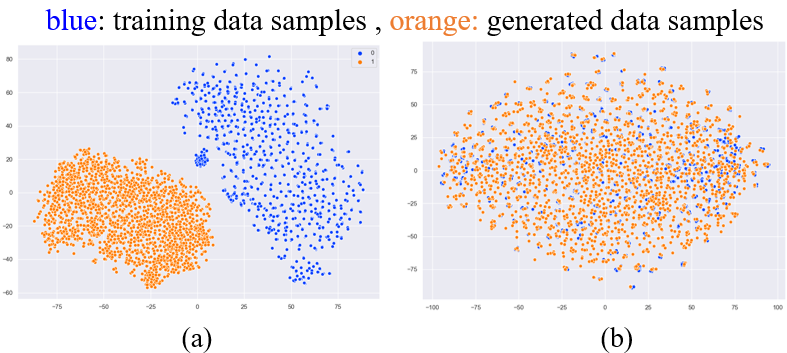}
    \caption{\textbf{t-SNE plots of training and generated samples.} (a) distribution of samples with registered scan data only, (b) distribution of samples after adding lifted data.}
    \label{fig:tsne}
\end{figure}

\noindent\textbf{Importance of training data generation}: After adding more data, we observed: i) less overfitting, ii) more diversity in the styles of the classes, and iii) higher fidelity. To quantify the improvement in fidelity and diversity with our training data generation, we used the t-SNE~\cite{tsne} plots which represent the image feature vectors as data sample points. Fig.~\ref{fig:tsne} demonstrates that adding more training data helps in moving the distribution of textures generated by ReAVAE closer to the distribution of training data (resulting in lower FID) compared to using only registered scans for training. We further fit an ellipse to each distribution and calculate the area under the ellipse. For the same number of data points, the generated samples in Fig.~\ref{fig:tsne}b occupy a larger area (area=2948.07) (\ie have larger diversity) compared to Fig.~\ref{fig:tsne}a (area = 2579.62). Additionally, in order to evaluate the improvement of a third-party image synthesis task with our generated textures, we trained StyleGAN2~\cite{stylegan2} from scratch with (a) our original training set and (b) equal number of textures randomly generated by our trained ReAVAE. We observed that the FID score of the output textures improved from 6.17 using (a) to 4.89 using (b), indicating that our generated textures exhibit more diversity than the training data while maintaining the fidelity.

\noindent\textbf{Importance of different components of ReAVAE}: Table~\ref{tab:ablation} measures the importance of different components in our network architecture. The baseline consists of the encoder-decoder architecture of~\cite{SEAN}. We observe that adding each part, as well as the alternate training strategy, gradually improves the results. The render loss, in addition to adding more constraints to the texture maps, ensures that the renders look seamless and captures details like crisp boundaries, folds and wrinkles. We also observed that instead of adding two more layers to our generator to produce a higher resolution image, generating an image at a lower resolution and then upscaling it using a super-resolution network is beneficial. This is because it is hard for normalization parameters at higher resolutions to distill meaningful information to the respective layers.

\begin{table}[t]
    \centering
    \caption{Quantitative evaluation of contribution of individual components of our network architecture.}
    \small
	\resizebox{0.95\columnwidth}{!}{
        \begin{tabular}{l cccc}
            \toprule
             \textbf{Method} & \textbf{PSNR}\(\uparrow\) & \textbf{SSIM}\(\uparrow\) & \textbf{FID}\(\downarrow\) &    \textbf{KID}\(\downarrow\)\\
                \midrule
                Baseline            & 16.53 & 0.68 & 32.15 & 0.024 \\
                + skip generator    & 17.42 & 0.71 & 31.19 & 0.022 \\
                + VAE               & 18.38 & 0.74 & 30.87 & 0.019 \\
                + training str.(final)     & \textbf{19.67} & \textbf{0.79} & \textbf{29.54} & \textbf{0.015} \\
                \hline
                w/ render losses    & 19.67 & 0.79 & 29.54 & 0.015 \\
                w/o render losses   & 18.65 & 0.75 & 30.75 & 0.018 \\
                \hline
                w/ SRN              & 19.67 & 0.79 & 29.54 & 0.015 \\
                w/o SRN             & 16.79 & 0.62 & 31.97 & 0.026 \\
                \bottomrule
            \end{tabular}
        \label{tab:ablation}
    }
\end{table}
\subsection{Applications}\label{ssec:apps}
\begin{figure}[t]
    \centering
    \includegraphics[width=1.0\linewidth]{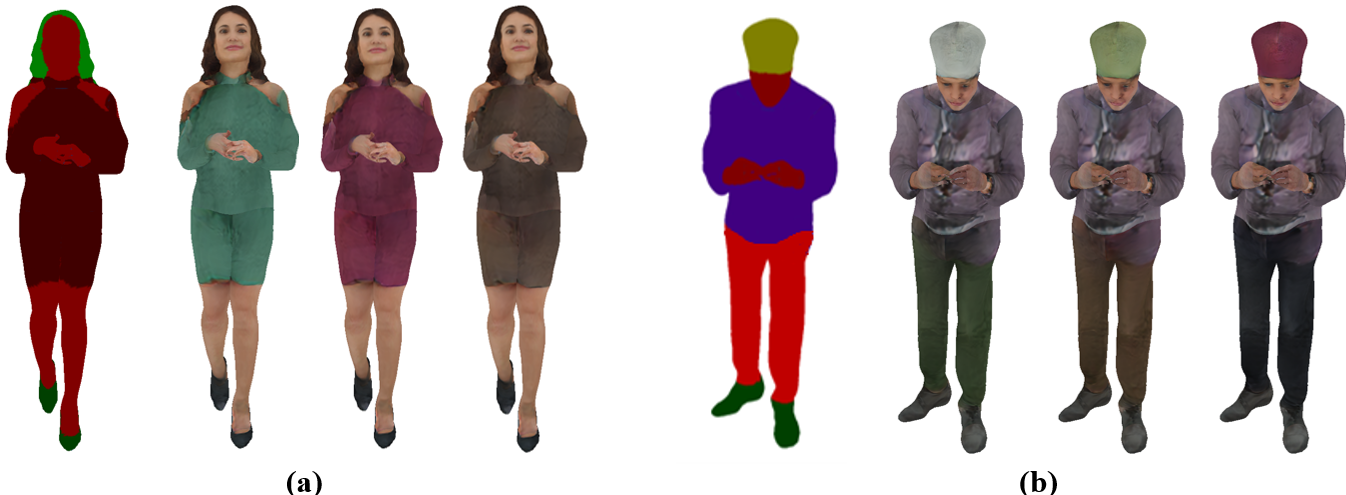}
    \caption{\textbf{Selective style editing}. (a) changing only the dress style, (b) changing only the headband and pants styles.}
    \label{fig:selectivechange}
\end{figure}

\begin{figure}[t]
    \centering
    \includegraphics[width=0.9\linewidth]{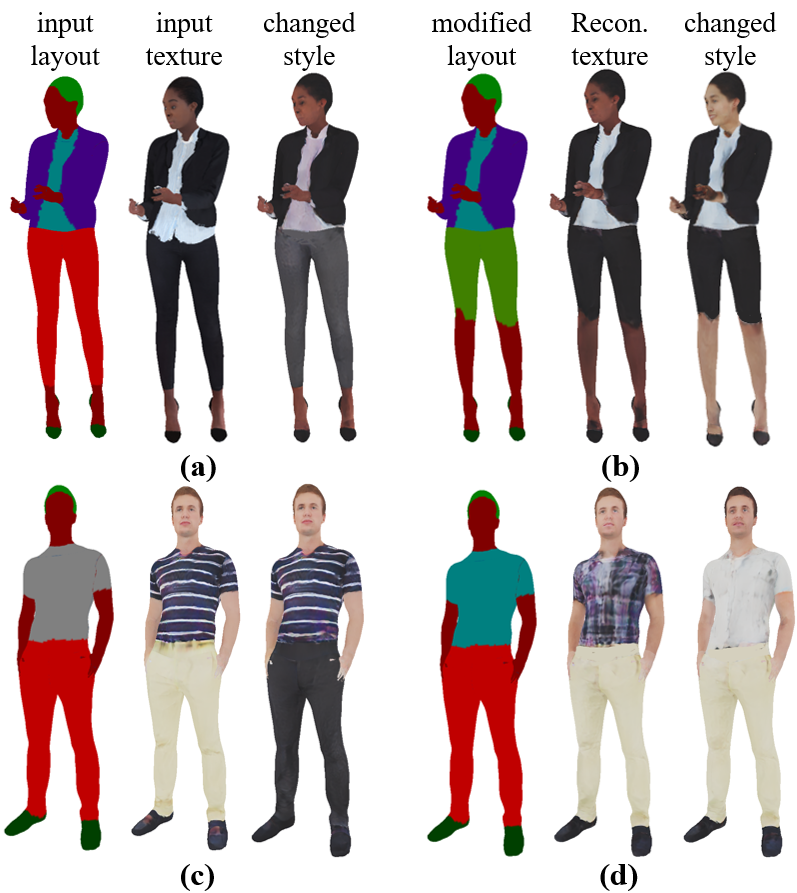}
    \caption{\textbf{Layout and style control}. We change the styles of (a) shirt and jeans, (b) skin, (c) pants, (d) shirt and hair. The reconstructed textures are generated by using all the styles from the input texture except for the class changed in the layout, for which an arbitrary style is used.}
    \label{fig:mixing}
\end{figure}

Our independent layout and style controllability together with the ability to perform either reconstruction or random synthesis enable us to generate a wide variety of textures with easy editability. Fig.~\ref{fig:mixing} shows two examples for independent layout and style editing. Layout editing includes changing jeans to shorts (\ref{fig:mixing}a and \ref{fig:mixing}b), t-shirts to shirts (\ref{fig:mixing}b and \ref{fig:mixing}c), short hair to long hair etc. Style editing enables changing the style of one or more classes at a time and mixing different styles for different classes. For example, in Fig.~\ref{fig:mixing}d, in the reconstructed texture we mix the styles of hair, skin, pants, and shoes from the input (exemplar) texture with a random style for the shirt. This example also shows that our method not only learns solid colors as garment styles but also checkered and other non-uniform patterns. We would like to refer the reader to the supplementary material for additional results.

Another example of selectively changing region styles in the random synthesis setup is given in Fig.~\ref{fig:selectivechange}. Here, we first apply random styles to all the classes and then fix all the styles except the ones we wish to change. Interestingly, although headband is a rare class (with limited examples in the training set), our network can generate vivid colors for this garment type without the need to search for an exemplar image with the desired headband color. 

\section{Conclusion}
We introduced a novel architecture that generates texture maps for 3D humans given an input segmentation mask in a semi-supervised setup. Our network uses a VAE to learn the per-class style distributions and enables controlling the generated texture by independently manipulating the layout through the mask and style by randomly sampling from the learned distributions. We demonstrated that our approach outperforms prior work in both the reconstruction and synthesis tasks and can be successfully applied in virtual try-on AR/VR applications. In the future, it will be interesting to synthesize the geometry and surface normals along with the textures for a complete unsupervised 3D avatar generation.

\noindent\textbf{Acknowledgements:} We thank Christoph Lassner, Olivier Maury, Yuanlu Xu and Ronald Mallet from Facebook Reality Labs for valuable discussions as well as the anonymous reviewers for their constructive feedback.

{\small
\bibliographystyle{ieee_fullname}
\bibliography{secRefs}
}

\clearpage
\section*{Appendix}
\begin{figure}[h]
    \centering
    \includegraphics[width=0.7\linewidth]{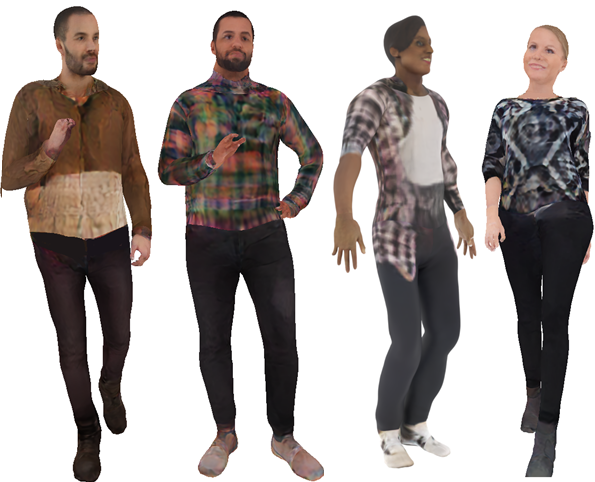}
    \caption{Examples of checkered and floral patterns in the textures synthesized by our proposed method.}
    \label{fig:complextextures}
\end{figure}

One of the main advantages of our method is the ability to perform texture synthesis under two different test scenarios: (a) synthesis guided by exemplar images and (b) synthesis with random styles. This provides the user with the flexibility to either copy a unique style from an exemplar image of choice or explore random styles outside the training distribution without the need to search for a corresponding exemplar image. In this supplementary material, we present more results of texture map generation by our method under these test scenarios. We also provide examples of complex patterns generated by our method in Fig.~\ref{fig:complextextures}, which shows that our network learns complex patterns in addition to uniform textures as garment styles. 

\section*{Synthesis guided by exemplar images}
Exemplar-guided texture synthesis can be further divided into three cases:
\begin{enumerate}
\itemsep 0em
    \item \textit{Reconstruction}, where the styles of all the regions of the output texture are taken from the input texture.
    \item \textit{Exemplar-guided style mixing}, where the styles from the input texture for some regions are mixed with the styles from an exemplar texture (different from the input texture) for the remaining regions.
    \item \textit{Non-exemplar guided style mixing}, where the styles from the input texture for some regions are mixed with the styles generated by random vectors for the remaining regions.
\end{enumerate}

\begin{figure*}
    \centering
    \includegraphics[width=0.625\linewidth]{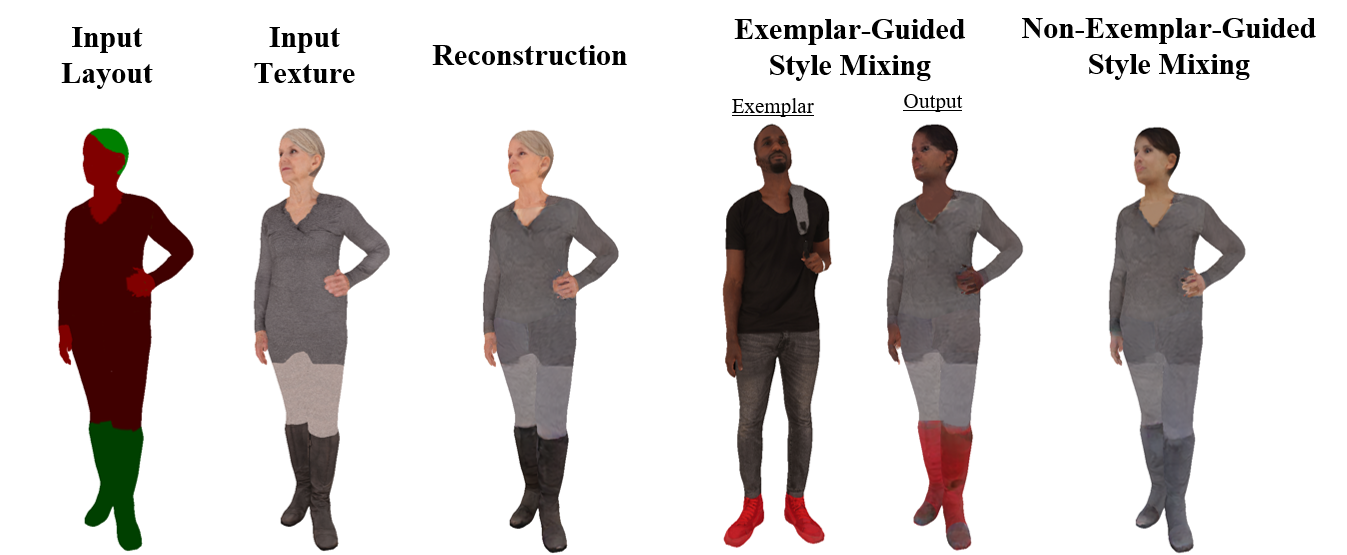}\\
    (a)\\
    \includegraphics[width=0.625\linewidth]{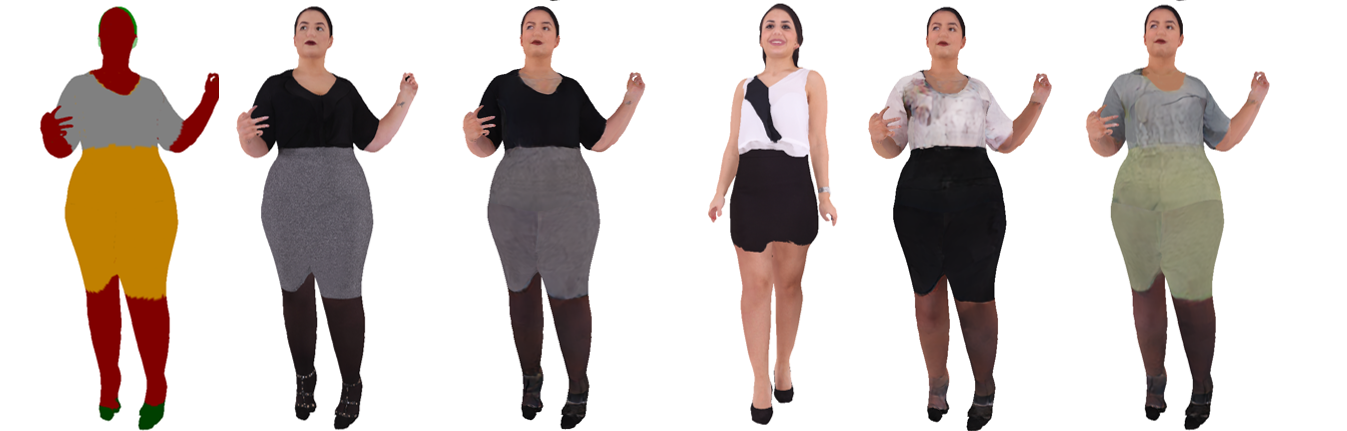}\\
    (b)\\
    \includegraphics[width=0.625\linewidth]{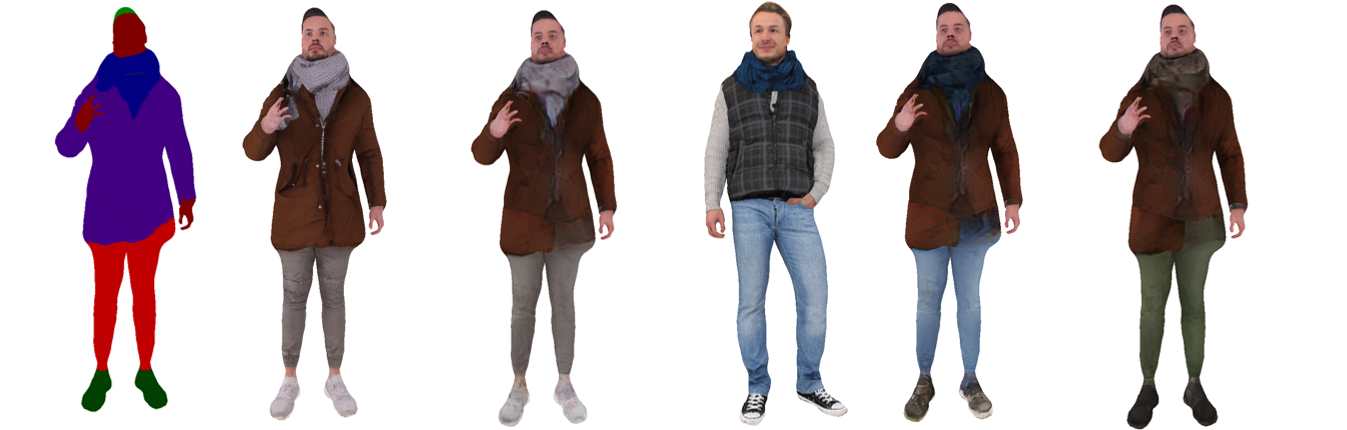}\\
    (c)\\
    \includegraphics[width=0.625\linewidth]{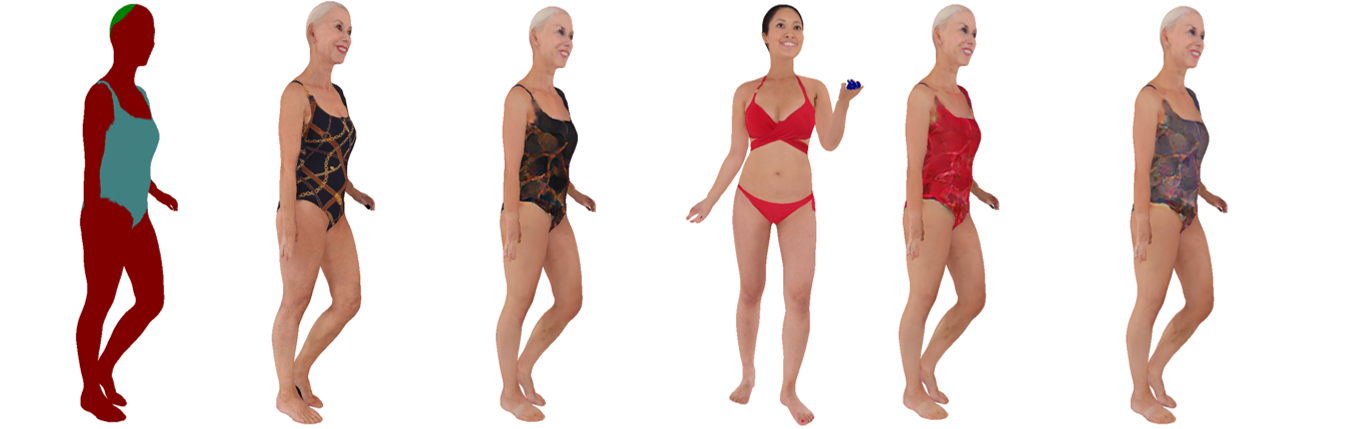}\\
    (d)\\
    \includegraphics[width=0.625\linewidth]{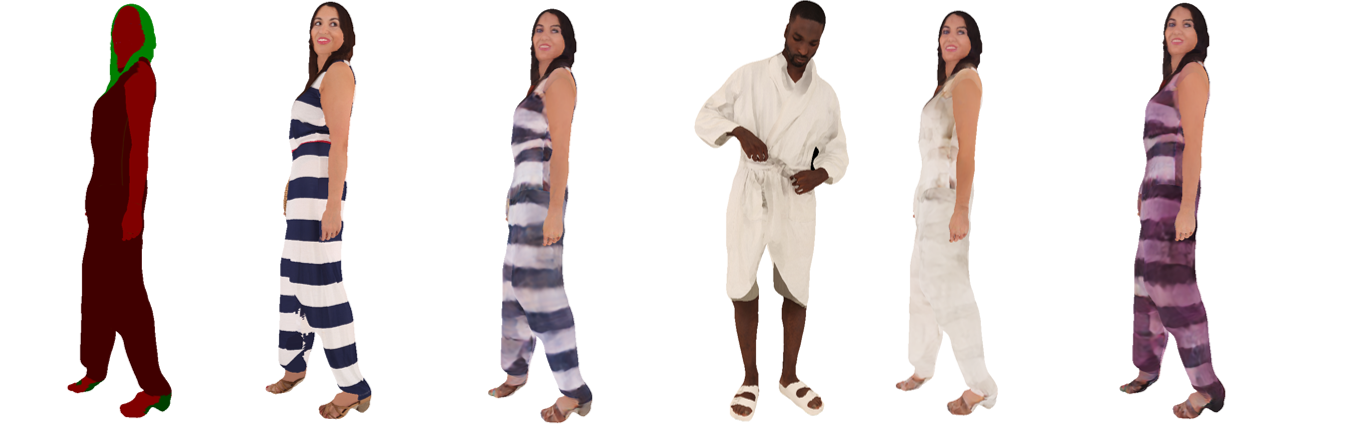}\\
    (e)
    \caption{\textbf{Synthesis guided by exemplar images}. We change the styles of the following classes: (a) hair, skin and shoes, (b) top and skirt, (c) scarf, jeans and shoes, (d) swimsuit, (e) jumpsuit and shoes.}
    \label{fig:exemplar}
\end{figure*}

\noindent Fig.~\ref{fig:exemplar} shows examples of textures generated with these three cases. We show examples with varying geometry, different garment types and shapes, and a variety of skin color and hair styles.

\section*{Synthesis with random styles}

Texture synthesis with random styles can be divided into two cases:
\begin{enumerate}
\itemsep 0em
    \item \textit{Random synthesis for all classes}, where we can randomly sample from the learned per-class style distributions to generate a random texture map.
    \item \textit{Style control of selected classes}, where we can fix the styles of some regions and change the styles of the remaining regions by controlling the corresponding style vectors in the style matrix at the generator's input.
\end{enumerate}

\begin{figure*}
    \centering
    \includegraphics[width=0.8\linewidth]{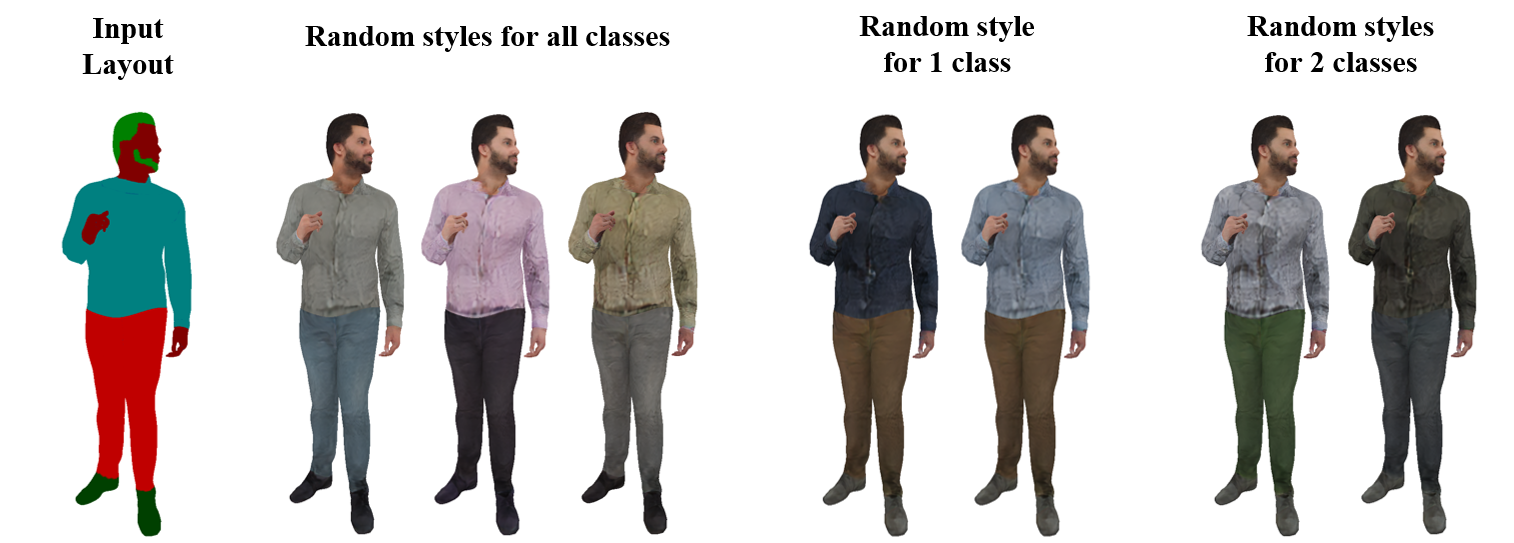}\\
    (a)\\
    \includegraphics[width=0.8\linewidth]{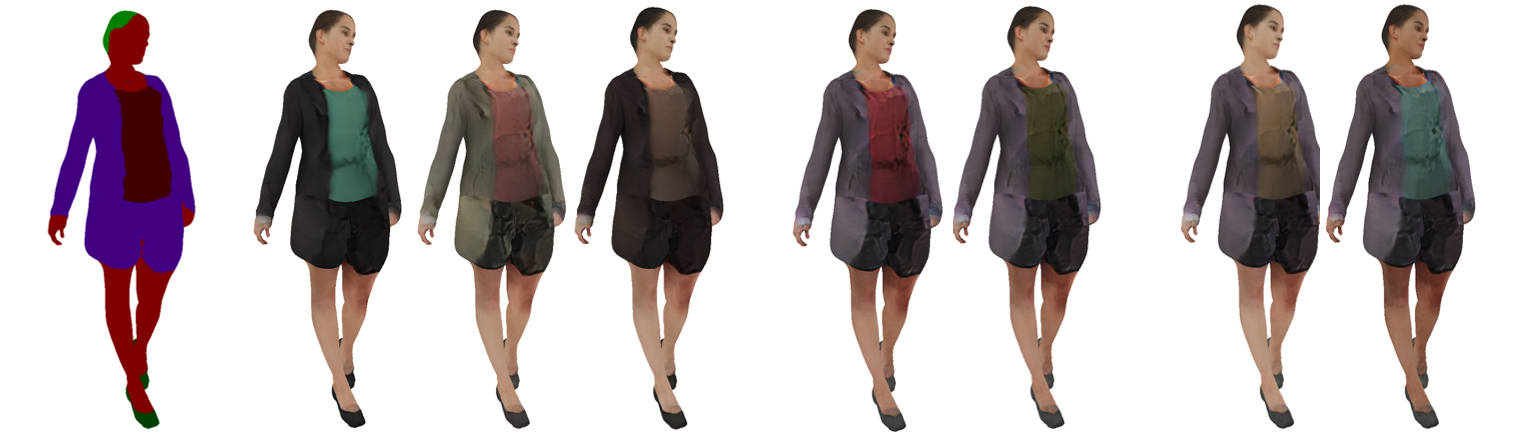}\\
    (b)\\
    \includegraphics[width=0.8\linewidth]{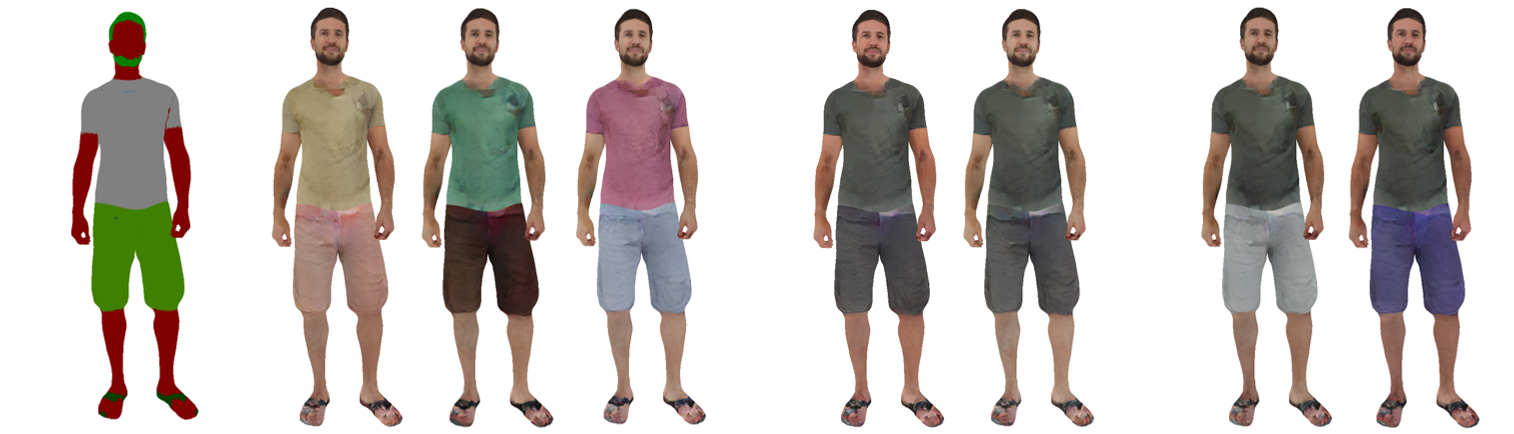}\\
    (c)\\
    \includegraphics[width=0.8\linewidth]{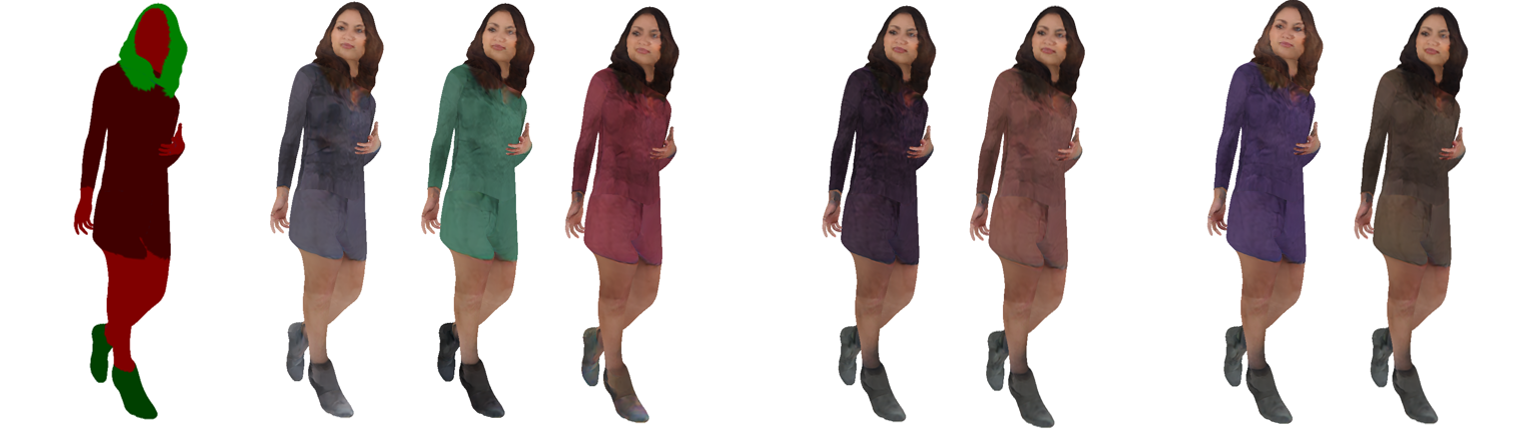}\\
    (d)
    \caption{\textbf{Synthesis with random styles}. The first 3 textures for each example are generated by sampling random styles for all the classes. The next 2 textures are generated by keeping the styles for all classes fixed except for a single class which is (a) shirt, (b) inner top, (c) skin, and (d) dress. The last 2 textures are generated by keeping the styles for all classes fixed except for two classes which are (a) shirt and pants, (b) inner top and skin, (c) skin and shorts, and (d) dress and hair.}
    \label{fig:nonexemplar}
\end{figure*}

We show examples to demonstrate these two test cases in Fig.~\ref{fig:nonexemplar}. Our network can successfully disentangle the contextual relationship among the styles of various regions. For example, although our training data contains examples of grey-haired persons wearing mainly suits and pants, our network can generate a grey-haired person wearing a t-shirt and shorts.

\section*{User interface for our application}

\begin{figure}[h]
    \centering
    \includegraphics[width=1.0\linewidth]{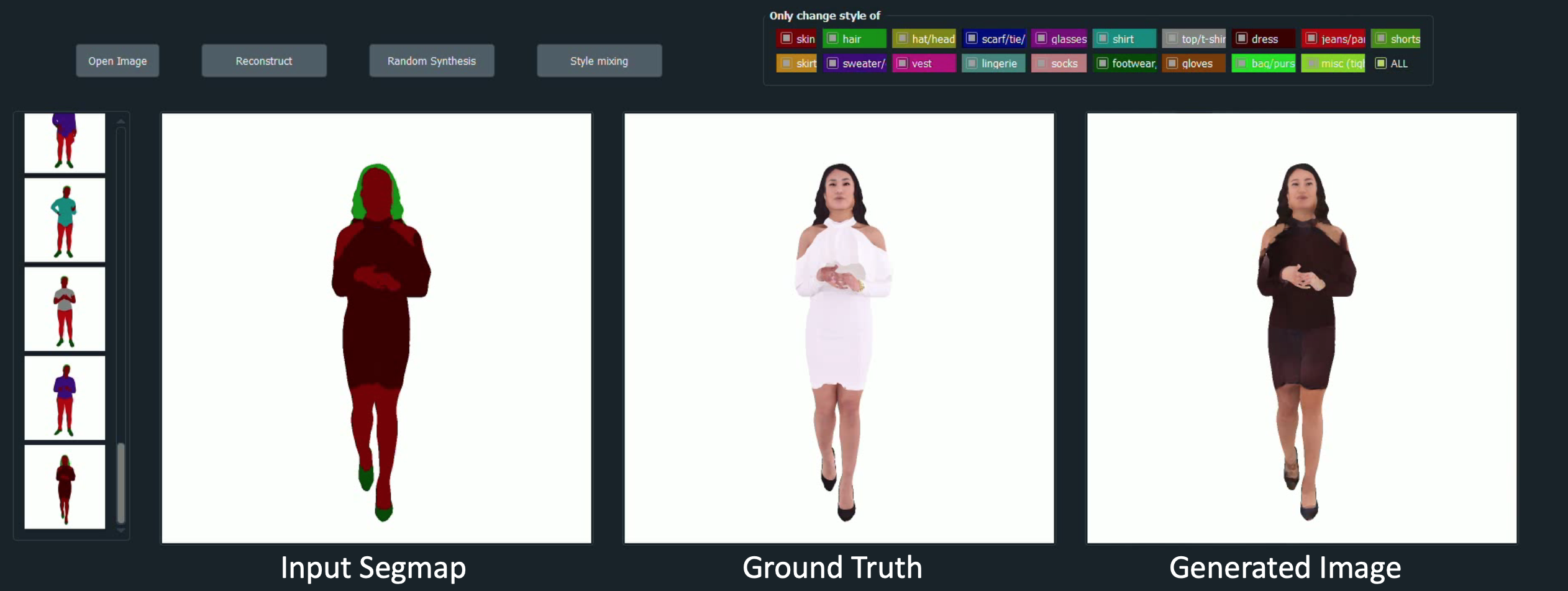}
    \caption{A screenshot of our designed user interface.}
    \label{fig:ui}
\end{figure}

We designed a user interface that allows a user to easily deploy our network for texture synthesis under the following scenarios: i) reconstruction of input exemplar image, ii) synthesis with random vectors, and iii) style mixing between exemplar styles and random styles (an example of which is shown in Fig.~\ref{fig:ui}). We would like to refer the reader to our webpage\footnote{\url{https://homes.cs.washington.edu/~bindita/humantexturesynthesis.html}} for additional forms of documentation of our method.

\end{document}